# Adapting LLM-SC to Living Sacred Heritage: Domain-Specific Extraction, Semantic Networks and Human Evaluation on Pashupatinath Temple Reviews

**Pujan Pant [1,*], Sagar Newpane [1], Hari Sitaula [2] and Manish Sakhakarmy [3,*]**

[1] Herald College Kathmandu, University of Wolverhampton, Naxal, Kathmandu 44600, Nepal
[2] Department of Mathematical Sciences, Montana Technological University, Butte, MT 59701, USA
[3] ASI Industrial, Billings, MT 59101, USA
[*] Correspondence: pantpujan017@gmail.com (P.P.); sakhakarmym@gmail.com (M.S.)

**Abstract:** Large language model (LLM) pipelines can turn visitor reviews into structured data, but generic schemes may miss rituals, sacred spaces, cultural restrictions and emotionally mixed experiences at living sacred heritage sites. We adapted the LLM–Social Computing (LLM-SC) framework to Pashupatinath Temple, Nepal, by specifying a 12-type entity taxonomy (four original LLM-SC types plus eight sacred-heritage types), four relation types and nine aspect-sentiment dimensions in a GPT-4o extraction prompt, which was applied to a fixed random sample of 2,500 English-language Tripadvisor reviews from a processed corpus of 3,939. The exports contain 11,286 entity, 6,232 aspect-sentiment and 5,590 relation records, covering 2,485, 2,487 and 2,433 reviews, respectively; the eight added sacred-heritage types account for 64.8% of entity records. Against an adjudicated two-annotator reference for 200 reviews, exact normalized surface-and-type matching gave provisional entity scores of 41.0% precision, 21.8% recall and 28.4% micro-F1. Aspect-presence F1 was 53.6% and end-to-end aspect-plus-polarity F1 was 42.6%. Human adjudication supported 153 of 194 first-listed relation candidates (one per review; 78.9%), a candidate-level rate that does not estimate relation recall or whole-graph precision. The contribution is a culturally grounded, domain-specific adaptation with component-wise human evaluation and explicit reproducibility and generalization limitations.

## 1. Introduction

Cultural heritage sites preserve historical, religious and cultural practices while continuing to shape contemporary visitor experiences. In Nepal, heritage is an important part of the tourism landscape, and both the Kathmandu Valley and Lumbini are inscribed on the UNESCO World Heritage List [1,2]. Tourism also contributes substantially to the national economy [3,4], while infrastructure and environmental problems make a systematic understanding of visitor experience increasingly important [3]. Online reviews are an increasingly important source for this purpose because they contain naturally occurring accounts of visitors' experiences, expectations, satisfaction and complaints [5–9].

Pashupatinath Temple lies within Pashupati, one of the seven Monument Zones of the Kathmandu Valley World Heritage property [1]. As a living sacred heritage site, it functions simultaneously as a place of worship, pilgrimage, cultural practice and tourism. Reviews of the site therefore contain more than conventional assessments of services or facilities: within a single account, visitors may describe rituals, religious actors, sacred spaces and objects, cultural restrictions, spiritual emotions, environmental conditions, crowding and interactions with site personnel. This creates a distinctive semantic challenge for automated review analysis, because a system must represent not only what visitors mention but also the cultural and experiential meaning attached to those concepts.

General-purpose review-analysis methods do not necessarily capture these characteristics. When culturally specific concepts such as rituals, religious actors, sacred places and objects, cultural rules and spiritual experiences are forced into broad generic categories, distinctions that matter to visitors can be lost. Sentiment poses a similar problem. A visitor may describe a strongly positive spiritual experience alongside crowding, pollution, restricted access or poor guidance, and a single sacred encounter may itself be described with both positive and negative emotions. A document-level polarity score cannot preserve these distinctions, and a generic aspect scheme may assign them to categories that do not reflect their sacred or experiential meaning.

Tourism and heritage research has increasingly applied natural language processing (NLP), transformer models and LLMs to visitor-generated content [5–13]. The LLM-SC framework [5] is a relevant foundation because it combines LLM-based semantic extraction, sentiment analysis and Social Computing network construction. However, a framework developed for one heritage context does not necessarily represent another adequately. When examined against Pashupatinath reviews, the original LLM-SC configuration did not provide sufficient representation for several culturally specific entities and experiential dimensions. The problem addressed here is therefore not simply how to analyze more reviews, but how to adapt LLM-SC so that it preserves the domain-specific semantic structure of living sacred heritage.

This study extends LLM-SC by adapting its semantic representation to living sacred heritage while retaining the framework’s core structure. The adaptation has two parts: a 12-type entity taxonomy that keeps four categories from the original configuration and adds eight sacred-heritage categories, and an explicitly guided extraction prompt that defines the entity types, four relation types, evidence requirements and nine aspect-level sentiment dimensions. The resulting structured information is represented as a Social Computing network of typed relations. GPT-4o outputs for a fixed sample of 2,500 reviews are evaluated against human-adjudicated reference annotations for a 200-review subset, separately for entity extraction, aspect-level sentiment and selected relation candidates. The evaluation asks whether the extracted content is supported by the review text, conforms to the schema and agrees with human judgments; a larger number of extracted records is not treated as evidence of better performance. Alternative models were screened only qualitatively, so no quantitative ranking of models is claimed. The structured outputs are motivated by recurring review monitoring, but hourly retrieval and live-system performance were not evaluated.

The study has three objectives:

1. To adapt the LLM-SC framework to living sacred heritage by developing a domain-specific entity taxonomy, extraction prompt and aspect-level sentiment representation capable of capturing culturally specific entities and complex visitor experiences;
2. To assess GPT-4o outputs against human-adjudicated annotations on a defined review subset, using component-appropriate measures for entity extraction, aspect presence and polarity, and relation support and type agreement;
3. To describe how the resulting structured outputs could support recurring visitor-review monitoring, while keeping this application-oriented motivation separate from any evaluation of live retrieval or system performance.

Section 2 reviews related work, Section 3 describes the data and methods, Section 4 reports the results, Section 5 discusses them together with the study’s limitations, and Section 6 concludes.

## 2. Related Work

### *2.1. User-Generated Content and Online Reviews as Data Sources for Tourism Research*

The growth of online platforms has made user-generated content (UGC) an important source for understanding visitor behavior, destination experiences and tourism performance. Fronzetti Colladon et al. [12], for example, analyzed more than 2.6 million Tripadvisor forum posts about seven European capital cities and found that forum language complexity and the centralization of the communication network were the strongest predictors of international airport arrivals, whereas models based on simple activity measures produced worse forecasts. Fei [7] similarly highlighted Tripadvisor, Instagram and X as complements to traditional surveys for capturing tourist experiences.

UGC is particularly valuable for cultural heritage because visitors describe their experiences in their own words rather than responding only to predefined survey dimensions. Research on historical districts in Fujian Province identified visitor priorities involving local culture, historical significance, architecture, food and commercial activity from online reviews [6], and Luo et al. [5] showed that visitor comments on the Forbidden City contain layered relationships among attractions, services and emotional responses that static survey instruments struggle to capture. For living sacred heritage, this naturalistic character is especially important, because reviews can contain spontaneous descriptions of rituals, religious actors, sacred spaces, spiritual experiences, access restrictions, environmental conditions and other factors that shape the visit.

The analytical value of UGC, however, depends on what can be recovered from the text. Because a single review covers several dimensions of experience, collecting large volumes of reviews does not by itself yield a structured understanding of visitor experience. Approaches are needed that identify the concepts visitors discuss, determine how those concepts are evaluated and represent their relationships within the wider heritage experience.

### *2.2. Sentiment Analysis and Aspect-Level Opinion Mining in Tourism*

Sentiment analysis is widely used to extract visitor opinions from tourism-related text. Early studies combined deep-learning sentiment classification with topic modeling to identify concerns such as local culture, historical significance, architecture, food and commercial activity in cultural heritage destinations [6]. Document-level sentiment is nevertheless a coarse representation, because a single review may praise some aspects of a visit and criticize others.

Aspect-level sentiment analysis addresses this by associating sentiment with particular dimensions of an experience. Fei [7] combined transformer-based language models and clustering to identify the tourism attributes associated with visitor satisfaction, including wellness services, natural environments and hospitality. Chen et al. [9] combined aspect-based sentiment analysis with machine learning and SHapley Additive exPlanations (SHAP) to identify concerns about overcrowding, accessibility and service quality at a heritage site. LLM-based approaches have also handled implicit and ambiguous opinions in hospitality reviews and approached human-level evaluation in some aspect-sentiment tasks [13].

These studies show the value of moving beyond a single positive or negative label. The distinction is particularly important for living sacred heritage, where a review may express strong spiritual satisfaction alongside dissatisfaction with access, crowding, environmental conditions or other operational factors, and where the same sacred experience may carry both positive and negative emotions. Aspect-level sentiment for sacred heritage therefore requires dimensions that reflect the domain rather than categories transferred unchanged from hospitality or general tourism.

### *2.3. Named Entity Recognition in Cultural and Heritage Domains*

Entity recognition determines which concepts in a review become available for later analysis. This matters in cultural and heritage settings because reviews contain domain-specific terminology, historical references, names and culturally meaningful expressions that may not map onto generic entity categories. Concepts that are not identified and represented correctly can direct downstream sentiment and relation analysis toward overly broad or semantically inappropriate categories.

Wang et al. [10] proposed GPT-NER, which reformulates named entity recognition (NER) as a text-generation task and performed strongly in low-resource settings where labeled training data are limited. Levchenko [11] found that GPT-4o recognized person names in Russian cultural-event announcements with higher F1 than transformer-based NER models. In cultural heritage tourism, Luo et al. [5] showed that LLM-driven extraction from Forbidden City reviews identified substantially richer tourism-related subcategories than a conventional NLP baseline.

Entity recognition is therefore more than a preprocessing step: it fixes the concepts available to downstream analysis. In living sacred heritage, the required entities extend beyond attractions, facilities and general tourism categories to religious practices, religious actors, sacred locations and objects, cultural restrictions, and spiritual or emotionally complex experiences. The task is to define an entity representation that reflects the semantic structure of the target domain, not simply to extract more entities. In the literature reviewed for this study, we found limited evidence of such a representation for Himalayan or South Asian living sacred heritage.

### *2.4. Large Language Models and Domain Adaptation in Tourism Text Analysis*

LLMs are increasingly applied to tourism text because they can interpret context, implicit meaning and domain-specific language without every extraction rule being encoded manually. Roumeliotis et al. [8] compared GPT-4o-family models, before and after fine-tuning, with a BERT baseline for hotel-review classification and reported higher accuracy for the GPT models. Água et al. [13] reported GPT-4o-based aspect-level sentiment results comparable to human analysis for hotel reviews. GPT-NER showed the potential of generative models for low-resource entity extraction [10], and Luo et al. [5] showed that LLM-based analysis gives greater semantic granularity on cultural heritage reviews than conventional rule- or dictionary-based approaches.

LLM-based approaches nonetheless remain sensitive to prompt design, computational cost, multilingual and metaphorical language, and the way the target domain is specified. Existing evaluations have often focused on hospitality, wellness or other settings where task-specific annotated resources are available [8,13]. For living sacred heritage, we did not identify a directly comparable public annotated resource for the culturally specific entity and experiential categories required here, which constrains conventional supervised domain adaptation.

Prompt-based domain adaptation offers an alternative under this constraint. Rather than assuming that a model's existing knowledge is sufficient for consistent downstream analysis, domain knowledge can be specified explicitly through entity definitions, examples, sentiment dimensions, relation types and extraction rules. The model's role is then not simply to recognize religious terms, but to map them consistently into a predefined semantic representation suitable for subsequent analysis. This distinction motivates the adaptation of LLM-SC developed here.

### *2.5. Social Computing and Semantic Network Analysis in Tourism Research*

Review-level extraction can be extended through Social Computing and network analysis, which model relationships among entities, behaviors and other elements of tourism systems. Fronzetti Colladon et al. [12] showed that network properties of tourism discussions carry information about online discourse and tourism demand, and Maghsoudi et al. [14] used social network analysis of 162 tourism-related websites in Iran to show that digital tourism services operate as an interconnected ecosystem.

Luo et al. [5] combined LLM-based semantic extraction with Social Computing to construct heterogeneous information networks from visitor reviews, representing relationships among attractions, services, visitor concerns and emotional responses. Such representations show how extracted concepts are connected within visitor narratives, which frequency counts or isolated sentiment scores cannot.

Network structure, however, depends on the quality and semantic specificity of the underlying extraction. A network built from generic noun phrases and undifferentiated co-occurrence differs from one in which entities carry domain-specific types and relationships are explicitly classified. For living sacred heritage, Social Computing is therefore best treated as a downstream component whose usefulness depends on the domain-adapted representation supplied by the extraction stage, not as an independent analytical layer.

### *2.6. Research Gaps and Positioning of the Present Study*

The studies reviewed above establish that online reviews provide rich evidence about visitor experience, that aspect-level sentiment captures dimensions obscured by document-level polarity, that LLMs can support entity extraction in low-resource settings, and that Social Computing can represent relationships among extracted concepts [5–14]. Three gaps remain for living sacred heritage.

The first is a representation gap. Tourism and heritage approaches commonly use general entity and sentiment categories, whereas living sacred heritage involves rituals, religious actors, sacred spaces and objects, cultural restrictions, spiritual emotions and emotionally complex experiences. The challenge is to represent the concepts visitors actually use, not merely to extract entities.

The second is an integration gap. Existing studies demonstrate entity extraction, aspect-level sentiment and network analysis, and some combine several of these. In the literature reviewed here, however, we found limited evidence of an integrated domain-adaptive LLM-SC framework that jointly specifies sacred-heritage entities, domain-specific sentiment dimensions and typed semantic relationships.

The third is an operationalization gap. Much of the existing literature evaluates approaches on retrospective datasets, and evidence on domain-adapted LLM-SC for ongoing review monitoring in living sacred heritage is limited. This study considers how its structured outputs could support monitoring, but its evaluation uses a fixed 2,500-review snapshot and does not test live retrieval or the reliability of an hourly monitoring service.

The contribution of this study is therefore a domain adaptation of LLM-SC for living sacred heritage, together with a component-wise evaluation of the selected GPT-4o pipeline against human-adjudicated reference annotations. The evidence does not establish that GPT-4o outperforms alternative models or that the adaptation transfers to other sites; rather, it provides an explicit adaptation procedure whose categories and definitions can be tested and refined in related sacred-heritage settings.

## 3. Materials and Methods

Figure 1 summarizes the workflow. Sections 3.1–3.3 describe corpus preparation, the domain-adaptive taxonomy and prompt, and the extraction pipeline; Section 3.4 describes the human-reference evaluation; and Section 3.5 describes semantic-network construction. GPT-4o served as the extraction model (Section 3.3); other generative AI tools were used to draft and refine selected passages of the manuscript, to prepare the plotting code for Figures 1 and 2 and to write analysis code for the evaluation statistics (Sections 3.4 and 4.2–4.5; see Acknowledgments).

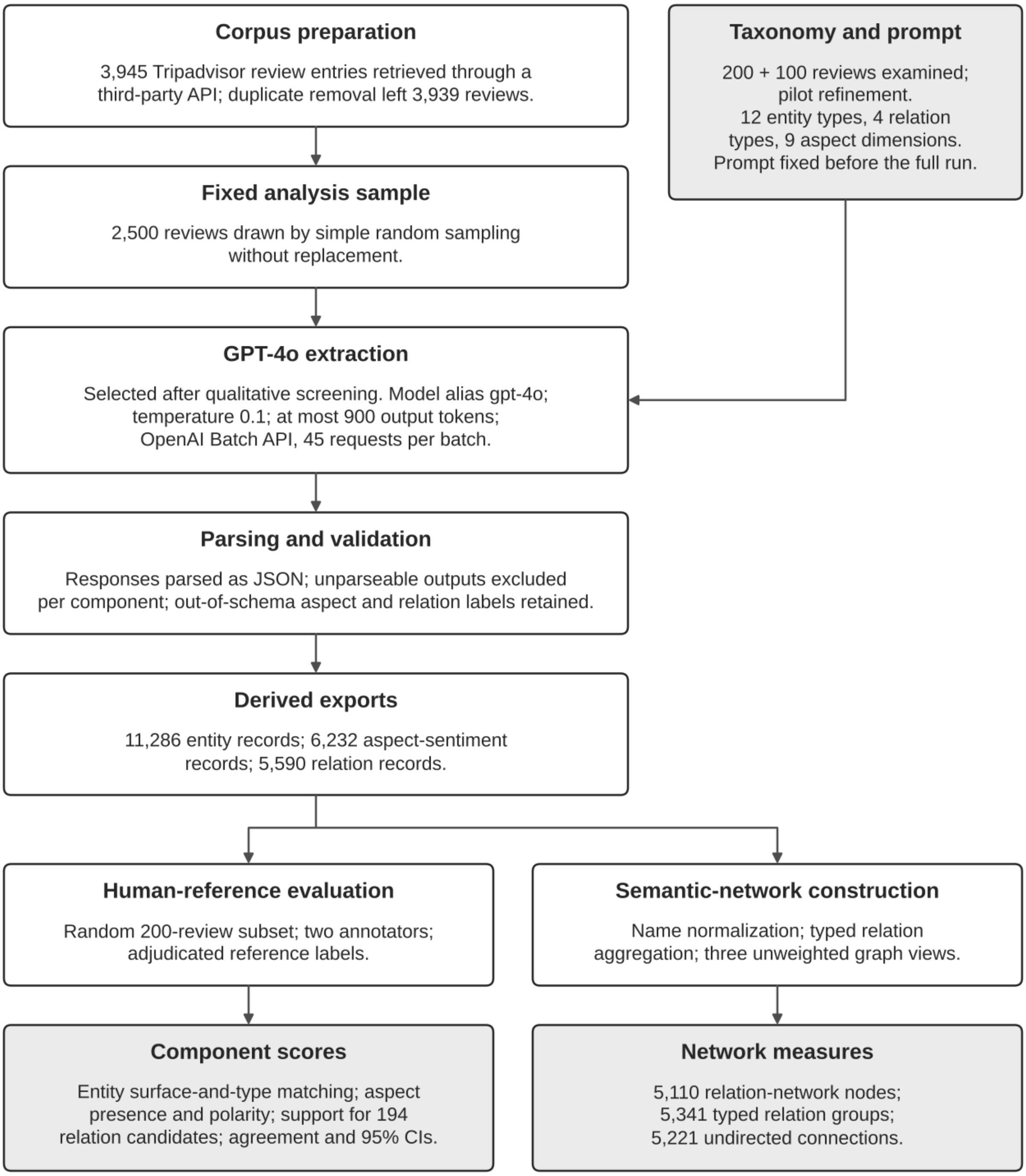

**Figure 1.** Workflow of the domain-adaptive LLM-SC pipeline and its two analysis branches. Shaded boxes show the external design input (taxonomy and prompt) and the reported outputs.

### *3.1. Data Collection, Preprocessing, and Sampling*

Reviews were retrieved from the Tripadvisor listing for Pashupatinath Temple, Kathmandu [15], through the Location Reviews endpoint of Omkar Cloud’s Travel Data API [16]. The endpoint returns structured review records with metadata, including review text, rating, date and trip type. Tripadvisor was selected because the preserved collection provided a comparatively large English-language corpus with useful metadata. Google Maps reviews were also accessible through the same service but were not used: preliminary inspection indicated that approximately 69% contained fewer than 10 words (median six words), too few for reliable entity and relation extraction.

The reviews were obtained from a third-party data provider rather than from Tripadvisor directly. Tripadvisor's Terms of Use restrict automated access to the site and the copying or redistribution of its content without written permission [17]. The authors did not obtain such permission and hold no documentation that the API provider was authorized to supply the reviews. This study therefore makes no claim of platform authorization or redistribution rights, and review texts are not shared (see the Data Availability Statement).

The initial collection snapshot contained 3,945 entries. Duplicate removal produced the processed corpus of 3,939 rows with 3,939 unique platform review IDs, covering reviews dated from 14 October 2008 to 16 April 2026. The six-entry reduction is reported to reflect duplicate removal, but the item-level exclusion log for this step is not available in the project files, so the record-level decisions cannot be reconstructed independently. Word counts, calculated from the word_count field of the processed corpus, have a mean of 67.01 (SD = 59.85), a median of 48 and a range of 8–718. These are descriptive statistics of the processed file; they do not establish which minimum-length threshold was applied at the raw-data stage.

Several auxiliary fields were derived for descriptive analysis. Publication dates were converted into year, month, quarter and broader period indicators, and trip type was flattened from the nested API field. A three-class sentiment_class variable was derived from the star rating (4–5 positive, 3 neutral, 1–2 negative); it served only as auxiliary metadata, not as the study's aspect-level sentiment output. A rule-based has_sacred_content indicator flagged reviews containing predefined lexical groups for rituals, religious actors, sacred spaces, spiritual concepts, festivals and access or cultural rules. These auxiliary variables were kept separate from the LLM-generated representations.

The fixed analysis sample contains 2,500 reviews from the 3,939-row processed corpus. Project records state that the sample was drawn by simple random sampling without replacement using Pandas. The preserved sample file is the authoritative input for all reported extraction results, and every sampled review ID and text field matches an entry in the processed corpus. The sample spans 2008–2026 and has a mean length of 68.34 words (SD = 60.04), a median of 49 and a maximum of 612; all sampled reviews are marked as English-language. The text_clean field of each sampled review was passed to the fixed extraction prompt (Section 3.2), review-level metadata were kept with the corresponding output records, and the same 2,500-review input file was used throughout. The text_clean field preserves letter case and ASCII punctuation; in the 200 evaluation reviews (Section 3.4) it contains no line breaks, repeated spaces, non-ASCII characters (such as emoji or typographic quotation marks), HTML codes or web links.

After the primary extraction and its main analyses had been preserved, a post hoc duplicate-handling sensitivity analysis was run on separate working copies. The audit identified five pairs of reviews with identical text under different platform review IDs. One record from each of four pairs was removed from the full-corpus sensitivity copy after comparison of the available metadata; both records of the fifth pair were retained because their reviewer labels and dates differed materially. A separate pair whose texts differed only in letter case was also retained and flagged. One of the four removed records belonged to the 2,500-review sample, so the sensitivity sample contains 2,499 reviews. No replacement review was drawn and no GPT-4o inference was rerun, so the sensitivity outputs are a post hoc filtered view of the original run rather than a new primary experiment.

### *3.2. Domain-Adaptive Taxonomy and Prompt Design*

The adaptation was implemented through a domain-specific representation scheme rather than by changing the underlying language model. Taxonomy development began with an examination of 200 randomly selected reviews from the eligible corpus to identify recurring concepts relevant to knowledge acquisition; a further 100 reviews were then examined to refine category boundaries and definitions. This review-informed process was complemented by domain knowledge about concepts that required explicit representation. The aim was to extend, not replace, the categories inherited from the original LLM-SC framework.

Four categories from the original configuration were retained (scenic_spot, problem, facility_service and general_sentiment, the last corresponding to the original emotions category) because they remained useful for architectural features, complaints, management and service experiences, and general emotional evaluations not specific to sacred practice. Eight categories were added for recurring sacred-heritage concepts, giving 12 entity types in total. The ritual category covers religious ceremonies and practices such as puja (worship involving offerings to a deity), aarti (a devotional ritual in which lamps are offered), darshan (beholding a deity image), cremation rituals and related acts. The religious_actor category covers people in religious roles, including sadhus (Hindu ascetic renunciants), priests, pilgrims, devotees, monks and yogis. The sacred_space category covers religiously significant locations, including ghats (stepped riverbank embankments used for ritual bathing and cremation), the Bagmati River, sanctums and cremation areas. The festival_event category covers named religious events, cultural_rule covers access norms, restrictions and culturally specific rules, and sacred_object covers physically identifiable religious objects such as offerings, lamps, garlands and ritual implements.

Two further categories represent aspects of sacred experience that general-purpose categories would struggle to preserve. The spiritual_emotion category captures an internal spiritual or emotional state explicitly experienced by the reviewer, rather than one attributed to the site or to another person. The dual_valence category captures an experience or entity that the reviewer explicitly describes with simultaneous positive and negative emotional meaning. It was introduced because sacred encounters can be emotionally mixed: a cremation, for example, may be described as disturbing or grief-inducing and, at the same time, as meaningful or transcendent, and treating it as simply positive or negative would discard part of what the review expresses.

Category definitions were refined during pilot inspection because some boundaries were initially less precise than consistent extraction required; definitions and examples were clarified to separate closely related concepts, particularly reviewer-felt spiritual emotion and mixed-valence experience. The final taxonomy and prompt were then fixed, and no prompt modifications were made in response to outputs from the 2,500-review extraction.

The prompt operationalized the taxonomy through natural-language definitions, examples and extraction rules (Appendix A). In addition to the 12 entity types, it specified four relation types (co_occurrence, causal, description and contrast) and nine aspect-level sentiment dimensions (service, facility, environment, ritual_experience, spiritual_authenticity, access_fairness, sacred_atmosphere, crowd_management and cultural_sensitivity). Exclusion rules were intended to prevent the site and generic contextual terms from being extracted repeatedly: Pashupatinath Temple, Kathmandu, Nepal and similar terms, and generic words such as place, site, visit and experience, were treated as contextual subjects rather than entities. The spiritual_emotion definition required the state to be attributable to the reviewer, and the dual_valence definition required explicit positive and negative responses to the same entity or experience; positive and negative statements about different aspects of a visit did not qualify. Relative to the original LLM-SC configuration, which used the Qwen-Max model with four entity types, three relation types (description, causes and co-occurrence), positive and negative polarity and a weighted co-occurrence–causal network [5], the adaptation adds eight entity types, a contrast relation, a neutral polarity class and nine named aspect dimensions, and analyzes unweighted graphs (Section 3.5).

### *3.3. LLM Extraction Pipeline*

#### *3.3.1. Model Screening and Extraction Configuration*

Several candidate approaches were screened qualitatively before the full run, including Llama-family models, a GLiNER-based extraction approach combined with SetFitABSA, and GPT-4-family models. Screening focused on tasks requiring contextual interpretation: domain-specific entity extraction, identification of reviewer-felt spiritual emotions and recognition of mixed-valence experiences. GPT-4o was selected on this basis. Because matched, reproducible outputs and reference-based evaluations are not available for the other candidates, the screening does not support a quantitative ranking, and quantitative claims in this study are limited to GPT-4o outputs evaluated against human references.

Each review's text_clean field was inserted into the fixed prompt (Appendix A) and submitted as a single user message through the OpenAI Batch API (Chat Completions endpoint), using the OpenAI Python library (August 2026 release), with the model alias gpt-4o, a temperature of 0.1, a maximum of 900 output tokens and 45 requests per batch. Requests were submitted in August 2026, and the prompt and request settings were held constant across batches. The dated model snapshot to which the gpt-4o alias resolved was not recorded.

#### *3.3.2. Structured Output Schema*

For each review, the model was instructed to return a single JSON object with five components (Appendix A): (1) entities, each with an identifier, an entity name taken from the review text, one of the 12 types and a supporting quotation of two to six words; (2) relations, each linking two entity identifiers with one of the four relation types and a brief description; (3) aspect-level sentiment records, each with one of the nine aspects, a polarity (positive, neutral or negative), a score and an evidence quotation; (4) an overall review type (devotional, experiential, critical or mixed); and (5) a short phrase describing the dominant experience. Scores range from 0.0 to 1.0. The prompt anchored 0.0–0.2 as strongly negative, 0.3–0.4 as mildly negative, 0.6–0.7 as mildly positive, 0.8–0.9 as strongly positive and 1.0 as superlative, and allowed exactly 0.5 only when the text contained equal positive and negative evidence for that aspect.

### *3.3.3. Batch Processing, Parsing, and Export Provenance*

A checkpoint file recorded the identifiers of processed reviews so that interrupted runs could resume without reprocessing. Each response was parsed as JSON; responses that could not be parsed were flagged as invalid and excluded from the corresponding component rather than corrected manually. In the aspect-sentiment and relation exports, labels outside the controlled vocabularies were retained: 11 aspect-sentiment records carry polarity labels outside the three classes (mixed, 10; dual_valence, 1), 7 carry aspect labels outside the nine dimensions, and 16 relation records carry the label cultural_rule. The entity export contains only the 12 taxonomy labels. In the exports, 2,485 of the 2,500 sampled reviews (99.4%) have at least one entity record, 2,487 (99.5%) at least one aspect-sentiment record and 2,433 (97.3%) at least one relation record; these are review-level coverage counts. The exports contain 11,286 entity records, 6,232 aspect-sentiment records and 5,590 relation records. No manual semantic correction of model-generated entities, relations or sentiment records was performed before scoring.

The entity export used for evaluation is a sanitized snapshot in which a third party's personal name is replaced by a redaction marker. The available post-processing code expands the parsed responses into entity, aspect-sentiment and relation tables but does not produce this evaluated file, so the entity export is treated as a fixed evaluation snapshot and the entity scores derived from it are reported as provisional (Section 4.3).

## ***3.4. Human Evaluation Design***

Two annotators, both authors of this study (Pujan Pant and Sagar Newpane), independently annotated a random subset of 200 reviews drawn from the 2,500-review sample, using a domain-specific annotation codebook. Agreement between the two annotations was measured before adjudication (Section 4.2), and disagreements were then reviewed and adjudicated to produce the reference labels. Entity agreement is reported as pairwise F1 between the two annotators' mention lists, because span annotation has no fixed set of negative cases; aspect and relation agreement are reported as raw agreement and Cohen's κ. For entities and aspects, the reference was constructed without inspecting GPT-4o predictions, to reduce circularity; relation adjudication, by design, judged selected GPT-4o candidates. Whether the reviews used for taxonomy development and prompt refinement (Section 3.2) overlap the evaluation subset was not tracked. One review in the subset has no model output in any component; it is scored as missing for entities and aspects and yields no relation candidate.

For entities, the adjudicated reference contains mention-level records, each with entity text, its character position in the review and one of the 12 taxonomy labels. Within each review, GPT-4o predictions are matched to reference mentions on normalized entity text (trimmed, case-folded and with internal whitespace collapsed) and entity type, treating both as multisets: for each review and each text–type pair, the number of true positives is the smaller of the predicted and reference counts. Unmatched predictions are false positives and unmatched reference mentions are false negatives. Because the prediction export provides no character offsets, these are string-and-type matching measures rather than character-boundary scores, and they are sensitive to surface-form variation and to the redaction policy.

For aspect-level sentiment, each review in the subset was annotated for all nine aspects, giving 1,800 review–aspect decisions. Aspect presence is evaluated separately from polarity. End-to-end scoring counts a prediction as correct only when the aspect is present in both the reference and the prediction and the polarities agree. Model labels outside the nine aspects are not counted as aspect predictions, and polarity labels outside the three classes are scored as errors. The nine decisions of the review without model output are retained as missing predictions and also counted as a coverage failure.

Relation evaluation examines source and target grounding, relation existence and type agreement for selected candidate relations. For each review in the subset that had relation output, the first relation in GPT-4o's output was selected as the candidate (194 candidates; six reviews had none), and all 194 candidate tuples were checked against the full relation export. Because this sample covers first-listed relations rather than a random sample of all relations, and does not annotate every possible entity pair, it supports candidate-level support and type-agreement rates but not full-output relation precision, recall or F1.

For entity matching, aspect presence and end-to-end aspect sentiment, counts of true positives (TP), false positives (FP) and false negatives (FN) are summed over all evaluated units before micro-averaged precision (P), recall (R) and F1 are computed:

$$P = \frac{TP}{TP + FP}, \quad R = \frac{TP}{TP + FN}, \tag{1}$$

$$F_1 = \frac{2TP}{2TP + FP + FN}. \tag{2}$$

Macro-F1 is the unweighted mean of the per-class F1 values over the $K$ classes in the set $C$ (the 12 entity types, or the three polarity classes):

$$F_1^{macro} = \frac{1}{K} \sum_{c \in C} F_{1,c}. \tag{3}$$

Conditional polarity accuracy is computed only over review–aspect pairs that both the reference and the model mark as present:

$$A_{pol} = \frac{N_{agree}}{N_{both}}, \tag{4}$$

where $N_{\text{both}}$ is the number of co-present pairs and $N_{\text{agree}}$ the number of those with identical polarity. For relations, the support rate $S$ and the typed-match rate $T$ are

$$S = \frac{N_{supp}}{N_{cand}}, \quad T = \frac{N_{typed}}{N_{cand}}, \tag{5}$$

where $N_{\text{cand}}$ is the number of selected candidates, $N_{\text{supp}}$ the number adjudicated as existing in the source review and $N_{\text{typed}}$ the number that are also assigned the same relation type as the final human label. For entities and aspects, missing outputs remain in the denominators; for relations, reviews without relation output yield no candidate and are reported as coverage failures (six reviews). Uncertainty is reported as 95% percentile bootstrap confidence intervals from 10,000 resamples of the 200 evaluation reviews, so that all mentions, decisions and candidates from a review are resampled together.

### *3.5. Semantic Network Construction and Analysis*

Entity and relation outputs were transformed into graph representations to examine how extracted concepts are connected. Records were normalized and aggregated in Python 3.12 with Pandas 3.0.5, and graphs were built and measured with NetworkX 3.6.1. Graph construction represents model-extracted relations; it does not verify that those relations are correct.

#### *3.5.1. Entity Normalization and Node Reconciliation*

Entity names and relation endpoints were normalized by trimming leading and trailing whitespace, case-folding and collapsing repeated internal whitespace, and the normalized string served as the node identifier. No lemmatization or synonym merging was applied, so inflectional and lexical variants (for example, cremation and cremations, or river and Bagmati River) form separate nodes. Observed name variants and assigned entity types were retained as node attributes. Where one normalized name had been assigned more than one entity type, the node was flagged as a type conflict rather than forced into a single type, preserving the ambiguity for later review.

A full entity inventory was kept separately from the relation network. Relation endpoints absent from the entity table were added to the inventory as relation-only nodes with an explicit unknown type, and extracted entities that never occur as a relation endpoint remain in the inventory but not in the relation-network topology. This keeps the inventory of all extracted concepts distinct from the subset connected by extracted relations.

#### *3.5.2. Typed Relation Aggregation and Direction*

Relation records were grouped by normalized source, normalized target, relation label and schema status. Because co_occurrence and contrast are semantically symmetric in this study, their endpoint order was canonicalized so that reversed versions of the same relation are not counted as separate groups; the source-to-target order of causal and description relations was preserved. Each group records the number of underlying relation records, the number of distinct reviews in which it occurs, the original labels and available text examples. Labels outside the four-type schema were retained and marked as out-of-schema rather than deleted or mapped to an expected type. Self-loop groups were kept in the typed representation for inspection but excluded from the topology measures described below.

#### *3.5.3. Graph Representations and Network Measures*

Three graph views were constructed. First, a typed multigraph retains separate relation groups between the same endpoints, with their labels, schema status, semantic direction, record counts and review counts; for co_occurrence and contrast the stored endpoint order is canonical only and carries no direction. Second, an unweighted undirected projection represents each distinct endpoint pair by one connection, combining relation labels and excluding self-loops; it summarizes overall topology. Third, a directional subgraph retains only causal and description relations; for its topology measures, parallel groups on the same ordered pair were combined and self-loops excluded.

For the undirected projection $G = (V, E)$ with $n = |V|$ nodes and $m = |E|$ connections, the average degree and density are

$$\langle k \rangle = \frac{2m}{n}, \quad D = \frac{2m}{n(n-1)}. \tag{6}$$

The local clustering coefficient of node $i$ with degree $k_i$ and $t_i$ triangles through it is

$$C_i = \frac{2t_i}{k_i(k_i - 1)}, \quad \langle C \rangle = \frac{1}{n}\sum_{i \in V} C_i, \tag{7}$$

with $C_i = 0$ when $k_i < 2$, so that the average clustering coefficient $\langle C \rangle$ includes every node. Transitivity is the global ratio

$$\tau = \frac{3T}{W}, \tag{8}$$

where $T$ is the number of triangles and $W$ the number of connected triples (paths of length two) in the projection. The projection is further described by degree centrality, the number of connected components and isolated nodes, the size and share of the largest connected component and the average shortest-path length within that component. The directional subgraph is described by its active nodes, unique directed connections, directed density, weakly and strongly connected components, its largest weakly connected component and in-degree and out-degree rankings. A sensitivity comparison repeated the undirected analysis after excluding out-of-schema relation groups. All measures were computed on unweighted graphs, and no edge-weight formula was applied. The measures describe the structure of the extracted representation; they do not establish real-world causal mechanisms, and highly connected nodes are structural hubs in the extracted network rather than independently validated causes or measures of visitor sentiment.

## 4. Results

### *4.1. Corpus-Level Extraction Coverage*

Table 1 summarizes component coverage for the 2,500-review sample. The entity export contains 11,286 records, the aspect-sentiment export 6,232 and the relation export 5,590. At least one record is available for 2,485 reviews in the entity export (99.4%), 2,487 in the aspect-sentiment export (99.5%) and 2,433 in the relation export (97.3%). These are component-specific coverage counts, not accuracy estimates, and relation-record counts are not graph-edge counts (Section 4.6).

**Table 1.** Component coverage for the 2,500-review analysis sample.

| Component | Reviews with output | Coverage (%) | Records in export |
|---|---|---|---|
| Entities | 2,485 | 99.4 | 11,286 |
| Aspect sentiment | 2,487 | 99.5 | 6,232 |
| Relations | 2,433 | 97.3 | 5,590 |

Coverage is the share of the 2,500 sampled reviews with at least one record of the component; it does not indicate correctness. The entity export is a sanitized snapshot (Section 3.3.3), and relations were validated at candidate level only (Section 4.5).

Table 2 shows how the entity records are distributed across the 12 types. The eight added sacred-heritage types account for 7,308 of the 11,286 records (64.8%); the most frequent added types are ritual (1,741 records) and spiritual_emotion (1,462), and dual_valence accounts for 220 records (1.9%).

**Table 2.** Entity records by taxonomy type in the entity export.

| Entity type | Origin | Records | Share (%) |
|---|---|---|---|
| ritual | Added | 1,741 | 15.4 |
| scenic_spot | Retained | 1,642 | 14.5 |
| spiritual_emotion | Added | 1,462 | 13.0 |
| sacred_space | Added | 1,425 | 12.6 |
| problem | Retained | 1,270 | 11.3 |
| cultural_rule | Added | 1,064 | 9.4 |
| facility_service | Retained | 876 | 7.8 |
| religious_actor | Added | 798 | 7.1 |
| sacred_object | Added | 493 | 4.4 |
| dual_valence | Added | 220 | 1.9 |
| general_sentiment | Retained | 190 | 1.7 |
| festival_event | Added | 105 | 0.9 |
| Total | | 11,286 | 100.0 |

Retained: category from the original LLM-SC configuration. Added: sacred-heritage category introduced in this study (together 7,308 records, 64.8%). Counts are output frequencies in the sanitized entity export (2,485 reviews), not accuracy estimates (Section 4.3).

### *4.2. Human-Reference Evaluation Sample*

The human evaluation uses the 200-review subset. Entity adjudication yielded 1,714 reference mentions. Aspect sentiment is represented as a complete grid of 1,800 review–aspect decisions (200 reviews × nine aspects), of which 509 are aspect-present and 1,291 aspect-absent. Relation adjudication also covers the 200 reviews: 194 had a GPT-4o relation candidate and six had none, including the review without any model output. The relation annotation is candidate-level rather than exhaustive over all extracted relations or possible entity pairs. The 199 subset reviews with model output did not differ significantly from the other 2,292 sampled reviews with output in star rating ($\chi^2$ test, $p = 0.16$), trip type ($\chi^2$ test, $p = 0.52$) or review year (Mann–Whitney test, $p = 0.77$; median 2016 in both), although the subset contains no one-star reviews, which make up 2.0% of the remainder.

Before adjudication, the two annotators marked 1,162 and 1,532 entity mentions. They agreed on 524 mentions with the same normalized span and type (pairwise F1 = 38.9%), rising to 778 (57.8%) when span variants confirmed by hand are included; where both annotators marked a span, they assigned the same type in 82.0% of cases (Cohen's κ = 0.80). For aspects, presence agreed in 91.9% of the 1,800 decisions (κ = 0.80), and polarity agreed in 82.3% of the 436 aspects that both annotators marked present (κ = 0.73). For relations, the annotators agreed on whether the candidate relation exists in 80.4% of the 194 candidates (κ = 0.38) and on its type in 67.4% of the 138 candidates both accepted (κ = 0.55). Chance-corrected agreement was lowest for relation existence (κ = 0.38).

### *4.3. Entity Extraction Against the Human Reference*

Under the exact normalized surface-and-type matching rule, the GPT-4o entity export yields 373 matched predictions among 910 predictions for the 200 reviews, leaving 537 unmatched predictions and 1,341 unmatched reference mentions (Table 3). Micro-precision is 41.0% (95% CI 37.7–44.4%), micro-recall 21.8% (19.5–24.1%) and micro-F1 28.4% (25.9–31.0%); macro-F1 across the 12 types is 26.5% (23.0–30.2%). These are mention-level surface-and-type scores, not character-offset span scores, because the export contains no predicted offsets. When entity type is ignored, 464 predictions match the text of a reference mention (precision 51.0%, recall 27.1%, F1 35.4%), and 80.4% of these surface matches also carry the reference type. Per-type F1 ranges from 49.4% for ritual to 0% for dual_valence (16 predictions and 3 reference mentions, none matched); general_sentiment reaches only 3.9% because the model rarely extracted it as an entity (13 predictions against 297 reference mentions; recall 2.0%).

**Table 3.** GPT-4o entity extraction against the human-adjudicated reference for 200 reviews (provisional values).

| Measure | Value (95% CI) | Definition |
|---|---:|---|
| Model predictions evaluated | 910 | Predictions for the 199 reference reviews with model output |
| Reference mentions | 1,714 | Adjudicated mentions with character positions |
| TP / unmatched predictions / unmatched reference | 373 / 537 / 1,341 | Exact normalized entity text and type, multiset-matched within review |
| Micro-precision | 41.0% (37.7–44.4) | 373/910 |
| Micro-recall | 21.8% (19.5–24.1) | 373/1,714 |
| Micro-F1 | 28.4% (25.9–31.0) | Equation (2) |
| Macro-F1 | 26.5% (23.0–30.2) | Unweighted mean over the 12 types (Equation (3)) |
| Type ignored: precision | 51.0% (47.7–54.3) | Exact normalized entity text only (464 matches) |
| Type ignored: recall | 27.1% (24.5–29.7) | 464/1,714 |
| Type ignored: F1 | 35.4% (32.6–38.1) | Equation (2) |

Intervals are 95% bootstrap confidence intervals (Section 3.4); the provisional status of these values is explained in Section 3.3.3.

The scores estimate agreement with the adjudicated reference under a strict text-and-type rule for this sample. An unmatched prediction is not necessarily an extraction error if the annotators omitted a valid mention, and the two predictions in which a redaction marker replaces a person's name count as unmatched. These values describe the evaluated snapshot and should not be read as definitive accuracy estimates for the full pipeline or for every review in the 2,500-review sample.

### *4.4. Aspect-Level Sentiment Evaluation*

Over the 1,800 review–aspect decisions, GPT-4o reaches a micro-precision of 54.8%, recall of 52.5% and F1 of 53.6% for aspect presence (Table 4). Where both the reference and the model mark an aspect as present, polarity agrees in 212 of 267 cases (79.4%). The stricter end-to-end evaluation of aspect and polarity together yields a precision of 43.5%, recall of 41.7% and F1 of 42.6%, and macro-F1 across the positive, neutral and negative classes is 41.6%. The two measures answer different questions: conditional polarity accuracy excludes disagreements about aspect presence, whereas the end-to-end measure also penalizes false-positive and missed aspects.

**Table 4.** GPT-4o aspect-presence and aspect-sentiment evaluation for 200 reviews.

| Measure | Value (95% CI) | Interpretation |
|---|---|---|
| Reference review–aspect decisions | 1,800 | 200 reviews × nine aspects |
| Reference present / absent | 509 / 1,291 | Adjudicated decisions |
| Aspect presence precision | 54.8% (50.6–59.0) | Micro-averaged over all 1,800 decisions |
| Aspect presence recall | 52.5% (48.6–56.4) | Micro-averaged over all 1,800 decisions |
| Aspect presence F1 | 53.6% (49.8–57.4) | Micro-averaged over all 1,800 decisions |
| Conditional polarity accuracy | 79.4% (74.6–84.0) | 212 of 267 co-present pairs (Equation (4)); not end-to-end performance |
| End-to-end precision | 43.5% (39.4–47.8) | Requires the correct aspect and polarity |
| End-to-end recall | 41.7% (37.7–45.7) | Requires the correct aspect and polarity |
| End-to-end F1 | 42.6% (38.7–46.6) | Requires the correct aspect and polarity |
| Macro-F1 (positive / neutral / negative) | 41.6% (37.4–45.7) | Noncanonical model labels are scored as errors |

Intervals are 95% bootstrap confidence intervals (Section 3.4). One review has no model output; 14 predicted evidence strings were not found in the review text even after normalization, and two predictions carry noncanonical polarity labels (see text).

An evidence check flagged 14 of the 487 in-scope aspect predictions because their evidence string was not found in the review after case, punctuation and spacing were normalized. These are quality-assurance flags rather than definitive hallucination labels, since the model may have paraphrased the source. Two predictions carry noncanonical polarity labels (mixed and dual_valence) and are scored as errors, and two rows with aspect labels outside the nine dimensions are not counted as aspect predictions. The flagged evidence strings remain in the scoring, and the review without model output is scored as missing (Section 3.4).

### *4.5. Semantic-Relation Validation*

All 194 selected GPT-4o candidate tuples match the full relation export on review ID, source node, target node and relation label. Of these, 153 were adjudicated as relations that exist in the source review, a sample-specific support rate of 78.9% (95% CI 73.1–84.5%; Table 5). A total of 106 candidates were both supported and assigned the same type as the final human label: 54.6% (47.7–61.5%) of all selected candidates and 69.3% (61.7–76.4%) of supported candidates. Across the 200 evaluation reviews, the export contains 450 relation records, but only one selected candidate per review was adjudicated, so these figures are not full-output precision, recall or F1 estimates.

**Table 5.** Validation of selected GPT-4o relation candidates, by model-assigned relation label.

| GPT-4o label | Candidates | Supported | Support rate (%) | Supported and type-matched |
|---|---|---|---|---|
| co_occurrence | 87 | 63 | 72.4 | 57 |
| causal | 32 | 28 | 87.5 | 13 |
| description | 43 | 38 | 88.4 | 23 |
| contrast | 29 | 22 | 75.9 | 13 |
| cultural_rule † | 3 | 2 | 66.7 | 0 |
| Total | 194 | 153 | 78.9 (73.1–84.5) | 106 |

† Label outside the four-type relation schema, retained rather than recoded. Supported and type-matched candidates have a positive final relation-existence decision and the same relation type as the final human label (Equation (5)); they make up 54.6% (47.7–61.5) of all candidates and 69.3% (61.7–76.4) of supported candidates. Intervals are 95% bootstrap confidence intervals (Section 3.4).

### *4.6. Semantic-Network Structure*

Graph counts depend on the unit counted, so Table 6 reports each count with its definition. The entity export contains 5,906 unique normalized entity names, of which 816 never occur as relation endpoints; the relation export adds 20 endpoint names absent from the entity table, which are kept as relation-only nodes of unknown type. The merged inventory therefore has 5,926 nodes, and the relation network has 5,110 distinct endpoint nodes. By model-assigned label, the 5,590 relation records comprise 1,962 co_occurrence, 1,420 description, 1,124 causal and 1,068 contrast records, plus 16 with the out-of-schema label cultural_rule. Type-aware aggregation of these records yields 5,341 typed edge groups: 5,325 with one of the four expected labels and 16 with the out-of-schema label cultural_rule.

The unweighted undirected projection has 5,221 connections among 5,110 nodes, an average degree of 2.043, a density of 0.000400, an average clustering coefficient of 0.01337 and a transitivity of 0.01583. It has 800 connected components and no isolated nodes; the largest component contains 3,269 nodes (63.97% of the relation network). Excluding the out-of-schema groups leaves 5,206 connections and 12 isolated nodes. The causal/description subgraph has 2,843 active nodes and 2,451 unique directed connections after parallel ordered pairs are combined and one self-loop is excluded. The value 5,308 is a separate diagnostic count of unique ordered source–target pairs ignoring relation labels, not a typed-edge count. All primary-run counts in Table 6 can be reproduced from the derived entity and relation exports using the rules in Section 3.5.

**Table 6.** Node and relation-graph counts and their definitions.

| Measure | Count | Definition |
|---|---|---|
| Unique normalized entity-name nodes | 5,906 | Distinct entity names in the sanitized entity export after trimming, case-folding and whitespace normalization |
| Relation-only endpoint nodes | 20 | Relation endpoints absent from the entity table; unknown type |
| Full merged node inventory | 5,926 | 5,906 entity-name nodes plus 20 relation-only endpoints |
| Nodes in relation network | 5,110 | Distinct normalized names occurring as a source or target |
| Relation records | 5,590 | Output rows across 2,433 reviews |
| Typed edge groups | 5,341 | 5,325 four-schema groups plus 16 out-of-schema cultural_rule groups |
| Unique ordered endpoint pairs, labels ignored | 5,308 | Diagnostic count only; not a typed-edge count |
| Undirected projection connections | 5,221 | Unique unordered endpoint pairs after collapsing labels and excluding self-loops |
| Causal/description directed connections | 2,451 | On the 2,843-node active subgraph; 2,471 typed groups before parallel pairs are merged |

Figure 2 shows every pair of concepts linked by at least seven relation records. It illustrates the model-generated structure; not every displayed edge was individually validated.

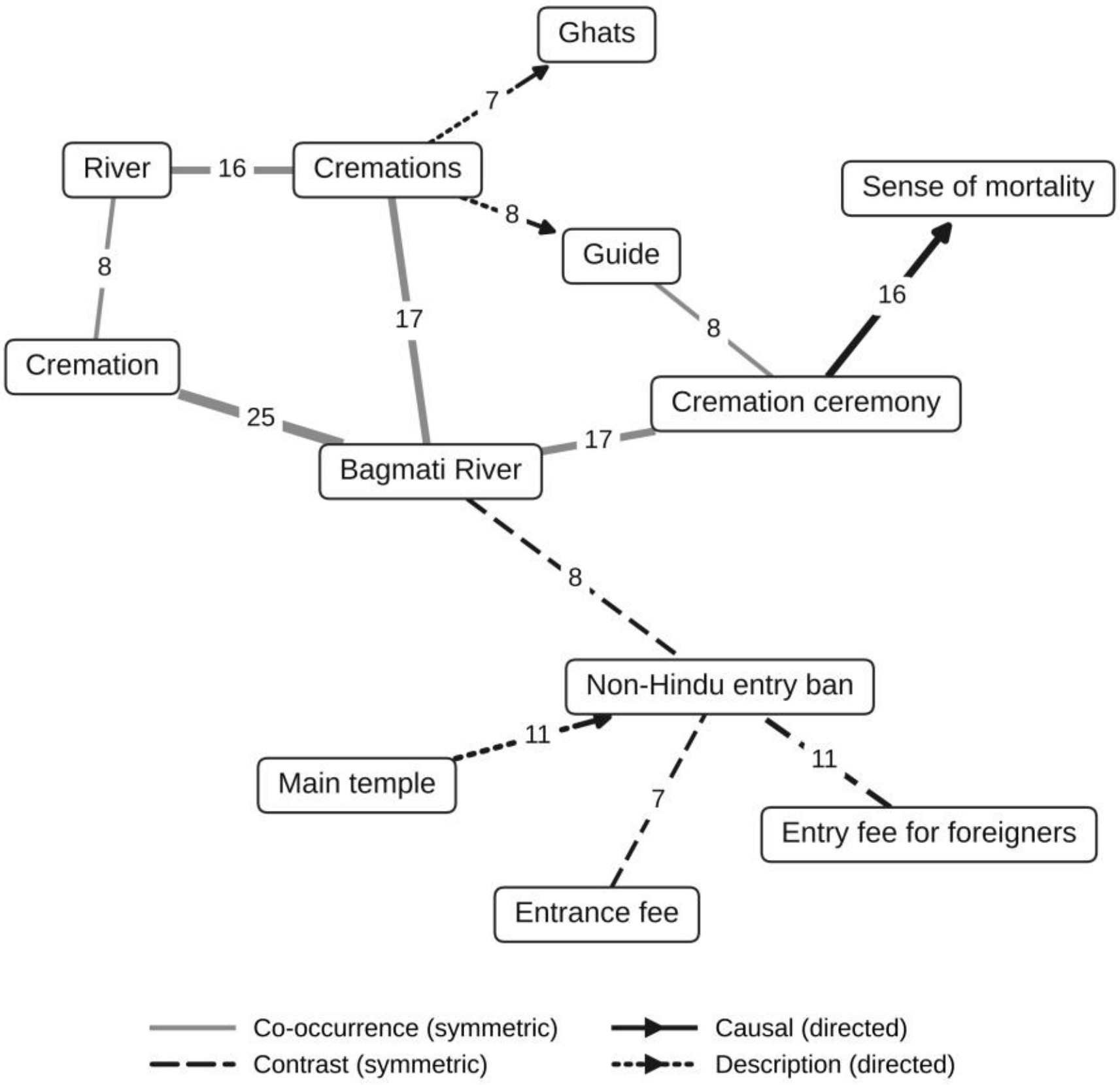

**Figure 2.** Every pair of concepts linked by at least seven relation records in the 5,590-record relation export (13 pairs among 12 nodes). Edge width and labels give the number of relation records for each pair (7–25), across all relation types and both directions. Each edge shows the most frequent relation type for its pair, distinguished by line style; causal and description edges point in the majority direction of their records, and co-occurrence and contrast edges are symmetric. Node names are normalized for case and whitespace only (Section 3.5.1) and shown with display capitalization; variants such as Cremation and Cremations therefore remain separate nodes. The displayed edges are model-generated and were not all individually validated.

### *4.7. Post Hoc Duplicate-Handling Sensitivity Analysis*

Sections 4.1–4.6 use the 3,939-review processed corpus and the original 2,500-review sample. To quantify the effect of the duplicate-handling decision, a separate sensitivity copy removed four redundant records from the full corpus and the one affected record from the sample and its model outputs, without replacement sampling or new inference. Table 7 compares the main corpus and output counts; its positive, neutral and negative classes are derived from star ratings, not from GPT-4o aspect-level sentiment.

**Table 7.** Corpus and component-output counts in the primary run and the duplicate-handling sensitivity copy.

| Measure | Primary run | Sensitivity copy | Change |
|---|---|---|---|
| Processed corpus reviews | 3,939 | 3,935 | −4 |
| Analysis-sample reviews | 2,500 | 2,499 | −1 |
| Full-corpus rating classes (positive/neutral/negative) | 3,409/347/183 | 3,406/346/183 | −3/−1/0 |
| Sample rating classes (positive/neutral/negative) | 2,156/229/115 | 2,155/229/115 | −1/0/0 |
| Entity output records | 11,286 | 11,282 | −4 |
| Reviews with at least one entity record | 2,485 | 2,484 | −1 |
| Aspect-sentiment output records | 6,232 | 6,230 | −2 |
| Reviews with at least one aspect-sentiment record | 2,487 | 2,486 | −1 |
| Mean exported aspect score (0–1) | 0.619031 | 0.618965 | −0.000066 |
| Relation output records | 5,590 | 5,588 | −2 |
| Reviews with at least one relation record | 2,433 | 2,432 | −1 |

At one decimal place, component coverage remains 99.4% for entities, 99.5% for aspect sentiment and 97.3% for relations when each numerator is divided by the corresponding sample size. These rates describe whether a review has output, not whether the output is correct. Table 8 compares the graph statistics.

**Table 8.** Graph statistics in the primary run and the duplicate-handling sensitivity copy.

| Graph measure | Primary run | Sensitivity copy | Change |
|---|---|---|---|
| Unique normalized entity-name nodes | 5,906 | 5,906 | 0 |
| Full merged entity inventory nodes | 5,926 | 5,926 | 0 |
| Nodes in the relation network | 5,110 | 5,110 | 0 |
| Typed edge groups | 5,341 | 5,340 | −1 |
| Expected-schema typed edge groups | 5,325 | 5,324 | −1 |
| Out-of-schema typed edge groups | 16 | 16 | 0 |
| Unique ordered source–target pairs, labels ignored | 5,308 | 5,307 | −1 |
| Undirected projection connections | 5,221 | 5,220 | −1 |
| Mean degree (undirected projection) | 2.04344 | 2.04305 | −0.00039 |
| Density (undirected projection) | 0.000399970 | 0.000399893 | −0.000000077 |
| Average clustering coefficient | 0.013370 | 0.013370 | 0 at six decimals |
| Connected components (all labels) | 800 | 801 | +1 |
| Nodes in largest component | 3,269 | 3,266 | −3 |
| Largest-component share of relation-network nodes | 63.97% | 63.91% | −0.06 percentage points |
| Average shortest path within largest component | 4.71361 | 4.70977 | −0.00384 |

The sensitivity copy leaves the merged entity inventory and the number of relation-network nodes unchanged. It reduces the typed-edge count, the unique ordered-pair count and the undirected connection count by one each; mean degree and density change slightly, and the average clustering coefficient is unchanged at six decimal places. Connectivity changes more: the number of connected components rises from 800 to 801 and the largest component loses three nodes. The duplicate-handling choice therefore has a small effect on most aggregate counts but can alter connectivity measures. The four removed review IDs do not appear in the human-reference workbooks, so the filtering does not affect the 200-review evaluation sample. The sensitivity results are reported separately and do not replace the primary results.

## 5. Discussion

### *5.1. Domain Adaptation and Semantic Representation*

The main contribution of the adaptation is the explicit representation of concepts that are salient in living sacred heritage but are not distinguished by generic entity labels. The eight added sacred-heritage categories account for 64.8% of the entity records in the export (Table 2), so the pipeline uses them extensively. Use is not accuracy or cultural validity, however: the categories must also be judged against the human reference, and both missed mentions and unsupported or mistyped mentions matter. The taxonomy is therefore best regarded as a domain-representation design whose performance is evaluated separately (Section 4.3), not as a taxonomy validated by the fact that the model populated it.

### *5.2. Separating Detection, Typing, and Polarity*

The component-level results show why output volume and a single aggregate score cannot characterize a structured-review pipeline. For entities, exact surface-and-type matching penalizes both lexical differences and taxonomy disagreement, and the absence of predicted offsets prevents boundary-level evaluation. Strict matching is also demanding for this task: before adjudication, the two annotators themselves agreed on the same normalized span and type for only 524 mentions (pairwise F1 = 38.9%). Output volume also limits recall directly: the evaluated export contains 910 predicted mentions for these reviews against 1,714 reference mentions, so even perfect matching could not have produced a recall above 53.1%. For sentiment, aspect-presence and end-to-end scores differ because a correct polarity cannot compensate for an omitted or spurious aspect. Conditional polarity accuracy (79.4%) is informative about the polarity step but should not be presented as overall sentiment performance. Neutral labels and noncanonical polarity outputs warrant targeted error analysis before stronger interpretations are drawn.

### *5.3. Interpreting Semantic Relations and Network Structure*

The semantic network represents model-extracted relations; it does not verify real-world causal mechanisms. In the relation validation, 153 of 194 selected relations (78.9%) were supported by human adjudication, and 106 (54.6%) were both supported and typed as adjudicated. These are support and typed-match rates for selected candidates, not exhaustive precision, recall or F1, because the 200 evaluation reviews contain 450 relation records, of which 194 were adjudicated. The graph analysis distinguishes the 5,926-node inventory, the 5,110-node relation network, the 5,341 typed groups and the 5,221-connection undirected projection (Table 6). Co-occurrence, description, causal and contrast relations have different meanings and should not be collapsed without an explicit reason, and symmetric relations require careful handling of endpoint order. Because names are normalized only for whitespace and case, variants of one concept (cremation and cremations in Figure 2) remain separate nodes, which spreads their connections across several nodes and can affect degree rankings and component counts.

### *5.4. Implications for Heritage-Review Monitoring*

A schema-guided pipeline could help analysts organize large volumes of visitor feedback into culturally specific entities, experiential dimensions and candidate relations. Potential uses include triaging recurring concerns about access, facilities, environmental conditions or cultural sensitivity, and identifying review narratives that merit human attention. These are potential decision-support uses rather than demonstrated management outcomes: the study evaluates a fixed snapshot and does not establish the reliability, latency or operational impact of a live monitoring service. Model-generated evidence should remain inspectable, and uncertain or high-impact interpretations should be reviewed by people familiar with the site and its cultural context.

### *5.5. Limitations*

Several limitations bound these findings:

1. *Data scope.* The corpus consists of English-language Tripadvisor reviews of one living sacred heritage site. Platform self-selection, language restriction, review-writing practices, repeated text across distinct platform IDs and the site’s cultural setting limit generalization to other destinations, languages, platforms and visitor populations;
2. *Evaluation scope.* The human reference covers a random subset of 200 of the 2,500 sampled reviews, so the component scores are estimates from that subset; the bootstrap intervals quantify review-level sampling uncertainty only. Both annotators are authors of the study, and overlap between the evaluation subset and the reviews used for taxonomy development and prompt refinement was not tracked;

3. *Entity matching.* The entity measure uses normalized surface-and-type matching because predicted offsets are unavailable; it penalizes lexical variation and redaction-induced mismatches and is not a character-span F1;
4. *Entity export provenance.* The available post-processing code does not reproduce the evaluated entity file from the raw model responses, so the lineage of the evaluated snapshot cannot be fully established and the entity scores are reported as provisional;
5. *Aspect evaluation.* The aspect-sentiment audit found evidence strings not found in the source even after case, punctuation and spacing normalization, noncanonical polarity labels, out-of-schema aspect labels and one review without model output. The automated evidence checks are review flags rather than adjudicated hallucination findings, but these cases can affect the estimates;
6. *Relation validation.* Only the first-listed relation of each review with relation output was judged (194 candidates in 200 reviews), which is not a random sample of the 450 relation records in those reviews, so the support rate does not estimate full-output precision, and recall and F1 cannot be estimated; the annotators agreed only fairly on relation existence ($\kappa = 0.38$) and moderately on relation type ($\kappa = 0.55$);
7. *Corpus deduplication.* The item-level log of the six removals from the 3,945-entry collection is unavailable. The post hoc sensitivity copy (3,935 corpus reviews and 2,499 sampled reviews, with no replacement sampling or new inference) is reported separately in Section 4.7 and does not resolve the missing log;
8. *Model version and reproducibility.* Requests used the gpt-4o alias, the dated snapshot it resolved to was not recorded, the model is closed, and the raw model responses, intermediate export code and evaluation files are not available as a single frozen release, so end-to-end reproducibility is partial. Neither the review texts nor the supporting quotations and evidence extracted from them are redistributed, so external readers cannot regenerate the model requests or inspect the evidence behind individual records;
9. *Comparison scope.* The study includes no matched quantitative comparison with alternative models or with the original LLM-SC configuration, and no evaluation of live monitoring. It supports a bounded assessment of the selected workflow, not a claim of model superiority or operational impact;

10. *Source authorization.* The reviews were obtained through a third-party API, and the authors hold no documentation that the provider was authorized by Tripadvisor, whose terms restrict automated retrieval and reuse of content [17]. No legal conclusion is drawn, and review texts are not redistributed.

### *5.6. Future Work*

Future comparisons should apply every model to the same reviews, schemas, normalization rules and scoring units. Entity assessment should report mention-level precision, recall, F1 and schema conformance; sentiment assessment should separate aspect detection from end-to-end polarity; and relation metrics should follow a probability-based or exhaustive candidate sampling design. Coverage and invalid or missing outputs should be reported alongside the scores. Replicated annotation, evaluation of relations beyond first-listed candidates and evaluation at related sacred-heritage sites are needed before broader claims can be made.

## 6. Conclusions

This study adapted the LLM-SC framework to reviews of a living sacred heritage site by integrating a sacred-heritage entity taxonomy, nine aspect-level sentiment dimensions and a typed relation schema in one structured extraction workflow. The human-reference evaluation shows why these components should be assessed separately, because entity surface-and-type matching, aspect detection, polarity classification and candidate-level relation validation reflect different sources of error. Entity extraction scores were low and provisional (28.4% micro-F1; 95% CI 25.9–31.0%), and aspect presence and end-to-end aspect sentiment reached F1 values of 53.6% and 42.6%. The relation validation supported 153 of 194 selected candidates (78.9%) and found 106 exact supported-and-typed matches, but its non-exhaustive sampling design does not support full-output relation precision, recall or F1. The graph analysis distinguishes records, typed relation groups, node inventories and topology connections, so the network summaries are reproducible from the released exports under explicit definitions. A post hoc duplicate-handling check changed output and edge counts only slightly, although it added one connected component.

For heritage managers, the potential value is a traceable decision-support resource that organizes candidate concepts and visitor experiences for human inspection while keeping uncertain evidence, schema violations and missing outputs visible. The study does not show that the representations are culturally complete or reliable enough for autonomous management decisions. Its conclusions are bounded by the incomplete lineage of the evaluated entity export, the missing deduplication log, the unavailable raw model responses and unrecorded model snapshot, the 200-review reference sample and the absence of a matched benchmark or live-system evaluation.

**Author Contributions**

Conceptualization, P.P.; methodology, P.P.; software, P.P. and S.N.; validation, S.N. and H.S.; formal analysis, P.P.; investigation, P.P.; data curation, P.P. and S.N.; writing—original draft preparation, P.P. and S.N.; writing—review and editing, P.P., H.S. and M.S.; visualization, P.P. and S.N.; supervision, H.S. and M.S.; project administration, M.S. All authors have read and agreed to the published version of the manuscript.

**Funding**

This research received no external funding.

**Institutional Review Board Statement**

Not applicable. This study analyzed publicly posted online reviews and involved no recruitment of, interaction with or intervention on human participants. The released outputs contain no reviewer usernames or profile information, and platform review identifiers were replaced with pseudonymous surrogate keys. No institutional ethics approval or exemption was obtained.

**Informed Consent Statement**

Not applicable. The study used publicly posted reviews; reviewers were not recruited or contacted.

**Data Availability Statement**

The derived entity, aspect-sentiment and relation records, the taxonomy specification, the extraction prompt and the analysis code are available at [https://github.com/pantpujan017/Pashupatinath-LLM-SC-]. The released records contain pseudonymous review keys, review metadata (year, period, trip type and star rating), entity names and types, aspect labels, polarities and scores, and relation endpoints and types; entity names and relation endpoints are short phrases taken from the reviews. The available code supports inspection of the derived outputs and the downstream analyses, but it does not enable complete end-to-end regeneration of the evaluated entity metrics or of all original model requests. The review texts were obtained from Tripadvisor through a third-party API (Section 3.1) and are not redistributed, because the authors have not established reuse or redistribution rights. For the same reason, the supporting quotations, aspect evidence and relation descriptions generated during extraction are not shared, and the released records are not represented as cleared for unrestricted redistribution. Restrictions therefore apply to the availability of these data.

**Acknowledgments**

During the preparation of this study, the authors used GPT-4o (OpenAI; API model alias gpt-4o; dated snapshot not recorded) as the extraction model described in Section 3.3, and ChatGPT (OpenAI; GPT-6 Luna), Claude Sonnet 5.5 and Claude Opus 5.5 (Anthropic) to draft and refine selected passages of the manuscript, to prepare the plotting code for Figures 1 and 2 and to write analysis code for the evaluation statistics (Sections 3.4 and 4.2–4.5). The authors have reviewed and edited the output and take full responsibility for the content of this publication.

**Conflicts of Interest**

The authors declare no conflicts of interest.

## Appendix A. Extraction Prompt

The prompt below was sent to GPT-4o as a single user message, with the cleaned review text inserted after the final line in place of [review text] (Section 3.3).

```
You are analyzing tourist reviews of Pashupatinath Temple, Kathmandu — a living UNESCO
sacred heritage site.
Extract entities, relationships, and aspect-level sentiments. Output STRICT JSON only — no
preamble, no explanation, no markdown fences.

ENTITY TYPES (use exactly these labels):
  "scenic_spot"       : Physical heritage structures, architectural features, monuments,
carvings, pagodas, spires, courtyards
  "problem"           : Complaints, negative experiences, management failures, pollution,
aggression, poor facilities
  "facility_service"  : Staff behavior, ticketing, guides, restrooms, food stalls, signage,
parking, entry process
  "general_sentiment" : Overall emotional tone not tied to any specific aspect
  "ritual"            : Religious ceremonies and acts: aarti, puja, cremation ceremony,
darshan, parikrama, tarpan, abhishek, havan
  "religious_actor"   : Specific people in religious roles: sadhu, priest, pujari, pilgrim,
devotee, monk, yogi — NOT generic tourists
  "sacred_space"      : Religiously significant locations: ghat, Bagmati River, inner
sanctum, cremation ground, lingam shrine, pagoda roof
  "spiritual_emotion" : Deep internal emotional/spiritual states explicitly felt by the
reviewer: awe, transcendence, devotion, reverence, peace, grief, humility, being moved,
overwhelmed, blessed, spiritual connection, sacred feeling, sense of mortality — MUST be
felt by the reviewer, not just described
  "festival_event"    : Named religious events: Maha Shivaratri, Teej festival, Bala
Chaturdashi, evening aarati ceremony
  "cultural_rule"     : Access norms and restrictions: non-Hindu entry ban, dress code
requirement, no photography zones, entry fee for foreigners
  "sacred_object"     : Religiously significant physical items: prasad, Shiva lingam,
incense sticks, flower offerings, trident (trishul), marigold garlands, diya lamp,
rudraksha
  "dual_valence"      : An experience the reviewer describes with SIMULTANEOUS positive AND
negative emotion — grief AND awe together, beautiful AND sad, moving AND disturbing IN THE
SAME SENTENCE about the SAME thing

RELATIONSHIP TYPES: co_occurrence | causal | description | contrast

SENTIMENT ASPECTS: service | facility | environment | ritual_experience |
spiritual_authenticity | access_fairness | sacred_atmosphere | crowd_management |
cultural_sensitivity

CRITICAL RULES — read carefully:
1. DO NOT extract "Pashupatinath Temple", "the temple", "Kathmandu", "Nepal", "India" as
entities — these are the review subject, not extractable entities
2. DO NOT extract generic nouns ("place", "site", "visit", "experience") as entities
3. spiritual_emotion MUST be an emotion the REVIEWER personally felt — phrases like "I was
moved", "I felt awe", "overwhelmed with emotions", "sense of peace", "humbled", "deeply
spiritual moment", "moved to tears" → ALL are spiritual_emotion. Extract the emotion name
as the entity ("awe", "sense of peace", "overwhelmed emotions"). Also classify as
spiritual_emotion: 'other worldly fascination', 'exquisite cultural experience', 'strangely
changed', 'sense of mortality' — these are reviewer-felt internal states."
4. dual_valence — STRICT definition with examples:
   ✓ CORRECT: "watching the cremation was both heartbreaking and transcendent" →
dual_valence entity = "cremation ceremony"
   ✓ CORRECT: "the funeral pyres were deeply sad yet strangely beautiful" → dual_valence
entity = "funeral pyres"
   ✗ WRONG: "dead bodies" alone → NOT dual_valence, just sacred_space or ritual
   ✗ WRONG: "monkeys", "crowd", "smoke" → NOT dual_valence
   ✗ WRONG: anything that is only negative or only positive → NOT dual_valence
   Only use dual_valence if the reviewer explicitly expresses BOTH a positive feeling AND a
negative feeling about the EXACT same thing
5. Sentiment scores — use the FULL range:
   0.0–0.2 = strongly negative ("terrible", "disgusting", "ruined")
   0.3–0.4 = mildly negative ("disappointing", "could be better")
   0.6–0.7 = mildly positive ("nice", "worth visiting", "good")
   0.8–0.9 = strongly positive ("amazing", "life-changing", "beautiful")
   1.0 = superlative ("best experience of my life", "absolutely perfect")
```

```
   NEVER use exactly 0.5 unless the text contains genuinely equal positive and negative
evidence for that specific aspect. NEVER use 0.0 unless the text is explicitly hateful or
disgusted.
6. "sacred_atmosphere" and "spiritual_authenticity" are ASPECT labels for sentiment
analysis — they are NOT entity types. If you want to extract an entity about the sacred
feeling or authenticity of a place, use "spiritual_emotion" instead.
7. sentiment value in aspect_sentiments MUST be exactly one of: "positive", "neutral", or
"negative" — no other values, no qualifiers like "mildly negative".
8. Return ONLY valid JSON — no markdown, no backticks, no extra text

OUTPUT FORMAT:
{
  "entities": [
    {"id": 1, "name": "<exact phrase from text, not generic>", "type": "<entity_type>",
"quote": "<2-6 word supporting quote from text>"}
  ],
  "relations": [
    {"from_id": 1, "to_id": 2, "type": "<relation_type>", "description": "<brief reason>"}
  ],
  "aspect_sentiments": [
    {"aspect": "<aspect>", "sentiment": "positive|neutral|negative", "score": 0.85,
"evidence": "<direct quote from review>"}
  ],
  "review_type": "devotional|experiential|critical|mixed",
  "dominant_experience": "<one phrase: what is this review primarily about>"
}

REVIEW:
[review text]
```